\documentclass{article}
\usepackage{spconf,amsmath,hyperref}
\usepackage{graphicx,subcaption} 
\usepackage{multirow,booktabs} 
\usepackage{enumitem} 
\usepackage{changepage} 

\title{Fine-Tuning Large Multimodal Models for Automatic Pronunciation Assessment}
\name{Ke Wang, Wenning Wei, Yan Deng, Lei He, Sheng Zhao}
\address{Microsoft, Beijing, China}
\begin{document}
%
\maketitle
\begin{abstract}
Automatic Pronunciation Assessment (APA) is critical for Computer-Assisted Language Learning (CALL), requiring evaluation across multiple granularities and aspects. Large Multimodal Models (LMMs) present new opportunities for APA, but their effectiveness in fine-grained assessment remains uncertain. This work investigates fine-tuning LMMs for APA using the Speechocean762 dataset and a private corpus. Fine-tuning significantly outperforms zero-shot settings and achieves competitive results on single-granularity tasks compared to public and commercial systems. The model performs well at word and sentence levels, while phoneme-level assessment remains challenging. We also observe that the Pearson Correlation Coefficient (PCC) reaches 0.9, whereas Spearman’s rank Correlation Coefficient (SCC) remains around 0.6, suggesting that SCC better reflects ordinal consistency. These findings highlight both the promise and limitations of LMMs for APA and point to future work on fine-grained modeling and rank-aware evaluation.
\end{abstract}
\begin{keywords}
Automatic pronunciation assessment, computer-assisted language learning, large language model,
large multimodal model
\end{keywords}
\section{Introduction}
\label{sec:intro}

Computer-Assisted Language Learning (CALL) provides an economical and scalable approach for language learners to improve multiple aspects of spoken language proficiency, including fluency, accuracy, prosody, intelligibility, grammar, and vocabulary~\cite{Kheir2023PA}. With increasing globalization and closer international collaboration, proficiency in foreign languages and strong oral communication skills have become essential. In this context, CALL systems play a pivotal role by delivering immediate and personalized feedback to support effective communication across diverse linguistic environments.

A typical CALL system includes components such as Mispronunciation Detection and Diagnosis (MDD), Automatic Pronunciation Assessment (APA), and Automated Essay Scoring (AES)~\cite{Ye2022Approach,Wu2024Prompting,Mao2019OR,Lin2020PA,Gong2022GOPT,Chao20233MH,Fu2024LLM,Wang2025Exploring}. MDD focuses on identifying phone-level errors by aligning recognized phone sequences with canonical references. Common approaches include the Goodness of Pronunciation (GOP) method~\cite{Witt2000Phone,Hu2015Improved} and Extended Recognition Network (ERN)~\cite{Qian2010Capturing,Li2016Tone}, as well as recent end-to-end models leveraging CNN-RNN-CTC architectures~\cite{Leung2019CNN} and self-supervised learning with wav2vec 2.0~\cite{Xu2021Explore}. While these methods provide detailed feedback, they primarily emphasize accuracy and often lack prosody and fluency evaluation~\cite{Gong2022GOPT,Wang2023Assessing}.

In contrast, APA aims at providing a holistic evaluation across multiple aspects (accuracy, fluency, prosody, completeness, stress and total) at different granularities, including phoneme, word, and sentence levels. Traditional APA systems primarily rely on ASR-derived features and statistical models, which often struggle to capture nuanced prosodic and fluency patterns. Recent research has advanced toward multi-granularity and multi-aspect frameworks~\cite{Lin2020PA,Gong2022GOPT,Chao20233MH,Cincarek2009PA,Do2023Hierarchical} and integrated MDD–APA models~\cite{Ryu2023Joint,Chao2025HMamba}, leveraging modern neural architectures such as Transformers and Mamba to enhance assessment accuracy and robustness.

Recent advances in Large Language Models (LLMs) and Large Multimodal Models (LMMs) have enabled strong language understanding and multimodal reasoning, driving their adoption in CALL tasks. Wang et al.~\cite{Wang2023Assessing} explored LLMs for phrase break evaluation, finding performance gaps compared to their proposed method. Other studies prompted LLMs with audio and text for MDD~\cite{Wu2024Prompting} and introduced multimodal LLMs for sentence-level accuracy and fluency scoring~\cite{Fu2024LLM}, achieving competitive results. Recent work~\cite{Wang2025Exploring,NatarajanBalaji2025Leveraging} also examined LLMs for APA and feedback generation. With LMMs, researchers have begun fine-tuning models for L2 oral proficiency assessment without additional audio adapters~\cite{Ma2025Assessment}, marking progress in overall score prediction.

Despite these advances, LMMs’ ability to handle multi-granularity and multi-aspect APA remains underexplored. APA requires not only content understanding but also extraction of speech features such as accuracy, fluency, and prosody~\cite{Wang2025Exploring}. Nearly at the same time with this study, LoRA-based~\cite{Hu2021LoRA} fine-tuning has been applied to Phi-4 multimodal models for APA and MDD~\cite{Ahn2025English}, but prior work focused only on sentence-level scoring. Moreover, the scarcity of APA-specific data limits LMM pretraining in this domain. To address these gaps, we investigate fine-tuning strategies for APA across multiple granularities and aspects using Qwen2-Audio-7B-Instruct~\cite{Chu2024Qwen2Audio}, hereafter referred to as Qwen2-Audio, chosen for its strong text–audio capabilities and instruction-following design for speech-related tasks. Our contributions are threefold:

\begin{itemize}[label=\textbullet, itemsep=0.3em, parsep=0.1em, topsep=0em, leftmargin=2em]
  \item We systematically evaluate the performance of fine-tuned LMMs on APA tasks in both multi-granularity and multi-aspect settings, as well as single-granularity and single-aspect settings.
  \item We identify and analyze the limitations of fine-tuned LMMs in phoneme-level assessment.
  \item We provide insights into evaluation metrics, highlighting the importance of rank-based measures such as Spearman’s rank Correlation Coefficient (SCC) for APA.
\end{itemize}

\section{Methodology}
\label{sec:method}

\begin{figure}[htbp]
\begin{minipage}[b]{1.0\linewidth}
  \centering
  \centerline{\includegraphics[width=1.0\linewidth]{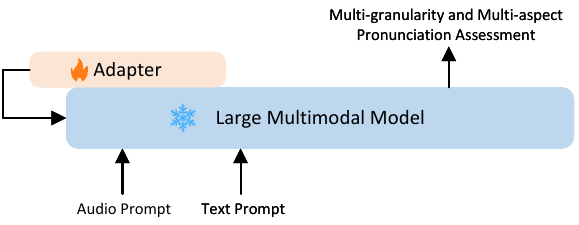}}
\end{minipage}
\caption{Overall structure of the proposed method.}
\label{fig:overall}
\end{figure}

Qwen2-Audio is capable of processing various types of audio inputs and operating in two distinct modes: audio analysis mode, which focuses on extracting and interpreting acoustic features, and voice chat mode, which generates direct textual responses based on spoken instructions. Due to its open-source availability and strong performance in audio-language tasks, we adopt Qwen2-Audio as the backbone for our APA framework. Unlike previous approaches that require an external audio encoder~\cite{Wu2024Prompting,Fu2024LLM}, Qwen2-Audio natively supports both audio and text inputs, significantly simplifying the pipeline and reducing system complexity for future research. The overall structure of the proposed framework is illustrated in Fig.~\ref{fig:overall}, and we employ LoRA for parameter-efficient fine-tuning of the LMM.

\subsection{Prompts}
\label{ssec:prompts}

Prompt engineering plays a critical role in adapting LMMs for APA tasks. We design multiple prompts tailored to different granularities and aspects of assessment. Below is an example of the comprehensive prompt used for training and evaluation, following the annotation guidelines:

\begin{adjustwidth}{-2.0em}{-2.0em}
\begin{quote}
\small
\itshape
You are a pronunciation evaluation teacher who analyzes speech at sentence, word and phone levels. At the sentence level, the evaluation includes accuracy, fluency, prosody, completeness, and a total score. Accuracy reflects phone-level pronunciation, accent, and clarity; fluency considers pauses, repetitions, and stammering; prosody assesses intonation, speed, and rhythm; and completeness indicates the percentage of words from the target text that were actually pronounced correctly. The total score is a comprehensive measure that reflects all aspects of speech. At the word level, the evaluation includes accuracy, stress, and a total score. Accuracy again considers phone pronunciation, accent, and clarity, while the stress score reflects the correctness of stress placement, where a score of 10 indicates correct stress and 5 indicates incorrect stress. At phone level, we assess the goodness of each phoneme with in the words. All scores range from 0 to 10, with 0 representing the poorest pronunciation and 10 representing the best, except for the word-level stress score, which is either 5 or 10.\textbackslash n Please evaluate the provided audio at the sentence, word, and phone levels using the given reference text and phone sequence.\textbackslash n Reference text: \{REFERENCE TEXT\}.\textbackslash n Reference phone sequence: \{REFERENCE PHONE SEQUENCE\}.\textbackslash n In the phone sequence, `-' indicates word boundaries. The final output format should be structured as follows:\textbackslash n Sentence Scores: \{Acc\} \{Flu\} \{Pro\} \{Com\} \{Tot\}\textbackslash n Word Scores: \{W1\}/\{Acc\}/\{Str\}/\{Tot\} \{W2\}/\{Acc\}/\{Str\}/\{Tot\} ...\textbackslash n Phone Scores: \{P1\}/\{Acc\} \{P2\}/\{Acc\} - \{P3\}/\{Acc\} ...~\footnote{To save space, we use the abbreviations `Acc', `Flu', `Pro', `Com', `Tol', `Str' to represent accuracy, fluency, prosody, completeness, total, and stress score, respectively, while `W' and `P' denote word and phone.}
\end{quote}
\end{adjustwidth}

For single-granularity or single-aspect tasks, as well as combinations of specific granularities and aspects, the prompt can be simplified according to the above instructions. Due to space limitations, we omit the actual simplified prompts in this section.

\subsection{Direct Preference Optimization}
\label{ssec:dpo}

Direct Preference Optimization (DPO)~\cite{Rafailov2023DPO} is a widely used method for aligning LLMs with human preferences through a classification-based objective over pairwise comparisons. While effective, DPO requires a reference model to compute relative rewards, adding memory and computational overhead. To address this, we adopt Simple Preference Optimization (SimPO)~\cite{Meng2024SimPO}, an efficient variant that removes the need for a reference model, improving scalability and generalization. The SimPO loss is defined as:
\begin{equation}
\mathcal{L}_{\text{SimPO}} = \log\!\Big(1 + e^{-\beta \big(r(y^+) - r(y^-) - \gamma\big)}\Big),
\end{equation}
\noindent where $r(y)$ is the average log-likelihood of the response $y$, computed as:
\begin{equation}
    r(y) = \frac{1}{|y|} \sum_{t=1}^{|y|} \log P(y_t \mid y_{<t}).
\end{equation}
\noindent Here, $r(y^+)$ and $r(y^-)$ denote reward scores for positive and negative samples, respectively. $\gamma$ is a reward margin, and $\beta$ controls the sharpness of the preference distribution. This formulation aligns optimization with the model's generation mechanism, improving efficiency and generalization. However, using only SimPO sometimes caused deviations from prompt instructions. To mitigate this, we combine SimPO loss with cross-entropy loss:
\begin{equation}
\mathcal{L} = \mathcal{L}_{\text{SimPO}} + \lambda\mathcal{L}_{\text{CE}},
\end{equation}
where $\lambda$ is a hyperparameter that balances the two components.

All SimPO training data is generated through simulation. Since Qwen2-Audio strictly follows prompt formats without shuffling word order, omitting sections, or removing word boundaries in phoneme scores, we only adjust the target aspect score. Positive and negative samples are created by randomly increasing or decreasing scores by 2 to 4 points. When phoneme scores are modified, the corresponding word and sentence scores are adjusted to maintain consistency. Additionally, if sentence-level scores such as prosody, accuracy, or fluency change, or if word-level scores such as accuracy or stress change, the corresponding total score is also updated.

\section{Experiments}
\label{sec:experiments}

In this study, we used two datasets to build and evaluate the proposed methods: a public dataset, Speechocean762~\cite{Zhang2021speechocean762} (referred to as SO762 for simplicity), and a private dataset. Both datasets were annotated by five independent language experts (LEs), and the final scores were obtained by averaging the annotations from all five LEs. Unlike SO762, our private dataset does not include phoneme-level annotations. To ensure annotation quality, we maintained inter-rater consistency: the Pearson Correlation Coefficient (PCC) and Spearman’s rank Correlation Coefficient (SCC) for sentence-level scores were required to exceed 0.6, while for word-level scores, PCC had to be greater than 0.6 and SCC greater than 0.5 between every pair of LEs.

\subsection{Datasets}
\label{ssec:data}

The SO762 dataset is an open-source and widely used resource for APA tasks. It comprises 5,000 English utterances produced by 250 native Chinese speakers. The training and test sets are randomly split, with each set containing 2,500 utterances from 125 speakers. The dataset provides detailed annotations for sentence-level metrics (accuracy, fluency, prosody, completeness, and total score) and word-level metrics (accuracy, stress, and total score). For phoneme-level annotations, the accuracy score originally ranges from 0 to 2, which differs from other scores that range from 0 to 10. For consistency, we converted phoneme-level scores to a linear scale of 0 to 10.

Our private dataset contains 20,410 English utterances collected from more than 500 English learners whose native language is Chinese, with roughly half being adults and half children. The annotation guidelines follow those of Speechocean762, and the dataset includes only sentence-level and word-level scores. We randomly split the data into 18,000 utterances for training and 2,410 for testing.

\subsection{Training and Evaluation Setup}

We trained and evaluated the model on a single NVIDIA GeForce RTX 4090 GPU (24 GB) using the bfloat16 floating-point format. The LoRA rank was set to 8 for all target modules. During training, the batch size was 1 with gradient accumulation steps of 8. Optimization was performed using AdamW with a cosine learning rate scheduler, allocating 10\% of the steps for warm-up to an initial learning rate of 1e-4. The LMMs were fine-tuned on the Speechocean762 dataset for 3 epochs and on the combined dataset (Speechocean762 plus our private dataset) for 2 epochs.  For SimPO training, $\beta$ was set to 0.1, $\gamma$ to 0.5, and $\lambda$ to 0.1. For evaluation, following~\cite{Wang2025Exploring}, we adopted PCC, SCC, and Root Mean Square Error (RMSE) as metrics.

\subsection{Multi-Granularity and Multi-Aspect APA}
\label{ssec:mgma}

\begin{table*}[th]
  \caption{PCC and SCC results for multi-granularity and multi-aspect APA on the SO762 dataset.}
  \label{tab:mgma_pcc_scc}
  \centering
  \resizebox{\linewidth}{!}{
  \begin{tabular}{l cc | ccc | ccccc}
    \bottomrule
    \multirow{2}{*}{Model} & \multicolumn{2}{c|}{Phoneme Score} & \multicolumn{3}{c|}{Word Score (PCC~/~SCC)} & \multicolumn{5}{c}{Utterance Score (PCC~/~SCC)} \\
    \cline{2-11}
    & RMSE & PCC/SCC & Accuracy & Stress & Total & Accuracy & Fluency & Prosody & Completeness & Total \\
    \hline
    GOPT~\cite{Gong2022GOPT} & 0.29 & 0.61~/~- & 0.53~/~- & 0.29~/~- & 0.55~/~- & 0.71~/~- & 0.75~/~- & 0.76~/~- & 0.16~/~- & 0.74~/~- \\
    HMamba~\cite{Chao2025HMamba} & 0.25 & 0.74~/~- & 0.71~/~- & 0.37~/~- & 0.72~/~- & 0.81~/~- & 0.85~/~- & 0.84~/~- & 0.28~/~- & 0.83~/~-  \\
    Azure PA~\cite{Wang2025Exploring} & - & - & 0.62~/~0.47 & - & - & 0.70~/~0.68 & 0.72~/~0.62 & 0.84~/~0.78 & 0.26~/~0.14 & 0.78~/~0.75 \\
    \hline
    FT & 0.39 & 0.38~/~0.34 & 0.51~/~0.46 & 0.11~/~0.11 & 0.52~/~0.46 & 0.69~/~0.63 & 0.74~/~0.70 & 0.73~/~0.67 & - & 0.72~/~0.67 \\
    SimPO & 0.39 & 0.38~/~0.34 & 0.52~/~0.47 & 0.08~/~0.08 & 0.53~/~0.47 & 0.68~/~0.62 & 0.73~/~0.69 & 0.73~/~0.68 & - & 0.72~/~0.66 \\
    \toprule
  \end{tabular}
  }
\end{table*}

We first evaluated the performance of LMMs fine-tuned for multi-granularity and multi-aspect tasks. The models were trained and assessed on sentence-level metrics (accuracy, fluency, prosody, completeness, and total score), word-level metrics (accuracy, stress, and total score), and phoneme-level accuracy. The experimental results are presented in Table~\ref{tab:mgma_pcc_scc}, where ``FT" denotes fine-tuning. The results indicate that fine-tuning the Qwen2-Audio model on SO762 achieved competitive performance at higher granularities but exhibited noticeable gaps in phoneme-level assessment. Incorporating SimPO with simulated data during training provided slight improvements in phoneme-level performance. Additionally, direct fine-tuning of Qwen2-Audio delivered a clear gain in sentence-level fluency compared with a commercial system (PCC: 0.74 vs. 0.72, SCC: 0.70 vs. 0.62), with SCC particularly highlighting improved ordinal consistency.

The completeness score consistently appears as \textit{NaN} due to the distribution of completeness scores in the training and evaluation data. Specifically, only 8 utterances in the 2,500-utterance training set have completeness scores ranging from 5 to 8, and only 14 utterances in the 2,500-utterance test set have completeness scores between 0 and 8.

Since the previous experiments revealed a clear gap in phoneme-level assessment, likely reflecting the limitations of LMMs in handling very fine-grained tasks such as phoneme-level scoring, we further explored performance at the word and sentence levels. To address the severe imbalance in completeness score distribution, we combined SO762 with our private dataset which has 1,872 utterances in the 18,000-utterance training set, and fine-tuned the model to examine whether this could further improve performance, particularly for completeness. The results are presented in Table~\ref{tab:ws_pcc}, where ``FT + Private" denotes training the model with both SO762 and our private dataset. Due to the highly skewed test data distribution, completeness score correlations remain unavailable even with a slightly more balanced training set. However, after adding additional training data, we observed significant PCC improvements at both the word and sentence levels, especially for the word-level stress score. Moreover, the experimental results also show that word-level accuracy, total score, and sentence-level prosody were significantly improved without predicting phoneme-level scores.

\begin{table}[thbp]
\caption{PCC results for word- and sentence-level scoring on the SO762 test set.}
\label{tab:ws_pcc}
\resizebox{\linewidth}{!}{
  \begin{tabular}{l ccc | cccc}
  \bottomrule
  \multirow{2}{*}{Model}  & \multicolumn{3}{c|}{Word Score} & \multicolumn{4}{c}{Sentence Score} \\
  \cline{2-8}
   & Acc & Str & Tol & Acc & Flu & Pro & Tol \\
  \hline
  FT & 0.57 & -0.01 & 0.58 & 0.69 & 0.74 & 0.78 & 0.72 \\
  SimPO & 0.58 & -0.01 & 0.60 & 0.69 & 0.74 & 0.73 & 0.72 \\
  FT + Private & \textbf{0.63} & \textbf{0.15} & \textbf{0.64} & \textbf{0.76} & \textbf{0.80} & \textbf{0.78} & \textbf{0.78} \\
  \toprule
  \end{tabular}
}
\end{table}

\subsection{Single-Granularity and Single-Aspect Scoring}
\label{ssec:sgsas}

Although previous experiments demonstrated performance improvements with reduced granularity, the upper bound for single-granularity and single-aspect evaluation remains unclear. To address this, we conducted five experiments: one on word-level accuracy and four on sentence-level aspects, including accuracy, fluency, prosody, and total score. The detailed results are presented in Table~\ref{tab:pcc_scc_results_sgsas}. These experiments reveal that zero-shot performance is extremely poor, whereas fine-tuning LMMs on single-granularity and single-aspect tasks yields significantly better results than multi-granularity and multi-aspect assessment, and in some cases, even surpasses certain public and commercial systems.

\begin{table}[htbp]
\centering
\caption{PCC and SCC results for single-granularity and single-aspect APA tasks.}
\label{tab:pcc_scc_results_sgsas}
\resizebox{\linewidth}{!}{
\begin{tabular}{l c  cccc}
\toprule
\multirow{2}{*}{Model} & \multirow{2}{*}{\shortstack[c]{Word Acc\\(PCC~/~SCC)}} & \multicolumn{4}{c}{Sentence Score (PCC~/~SCC)} \\
\cmidrule(lr){3-6}
 & & Acc & Flu & Pro & Tol \\
\midrule
GOPT & 0.61~/~- &    0.71~/~-    & 0.75~/~-    & 0.76~/~-    & 0.74~/~-    \\
Azure PA & 0.62~/~0.47 & 0.70~/~0.68 & 0.72~/~0.62 & \textbf{0.84}~/~\textbf{0.78} & \textbf{0.78}~/~\textbf{0.75} \\
\midrule
Zeor-Shot & -0.03~/~-0.02 & - & - & - & - \\
FT   & \textbf{0.62}~/~\textbf{0.57} & 0.74~/~0.69 & 0.79~/~0.78 & 0.77~/~0.76 & 0.77~/~0.71 \\
DPO  & 0.60~/~0.55 & \textbf{0.76}~/~\textbf{0.70} & \textbf{0.79}~/~\textbf{0.78} & 0.78~/~0.76 & 0.76~/~0.71 \\
\bottomrule
\end{tabular}
}
\end{table}

\subsection{Analysis of Correlation Metrics}
\label{ssec:analysis_pcc_scc}

The evaluation results on our private test set are shown in Table~\ref{tab:pcc_scc_results_so2024}. The PCC values are approximately 0.9, indicating a strong linear correlation. In contrast, the SCC values are around 0.6 across sentence-level metrics such as accuracy, fluency, prosody, and total score. This discrepancy suggests that SCC may serve as a more suitable evaluation metric for APA tasks as model quality improves, since PCC assumes a linear correlation and requires data to be continuous and approximately Gaussian distributed, whereas SCC only assumes ordinal relationships.

\begin{table}[htbp]
\centering
\caption{Evaluation results on the private test set}
\label{tab:pcc_scc_results_so2024}
\resizebox{\linewidth}{!}{
  \begin{tabular}{l ccc | ccccc}
  \bottomrule
  \multirow{2}{*}{Model}  & \multicolumn{3}{c|}{Word Score} & \multicolumn{5}{c}{Sentence Score} \\
  \cline{2-9}
  & Acc & Str & Tol & Acc & Flu & Pro & Com & Tol \\
  \hline
  PCC & 0.87 & 0.85 & 0.87 & 0.90 & 0.90 & 0.88 & 0.95 & 0.92 \\
  SCC & 0.74 & 0.82 & 0.75 & 0.62 & 0.59 & 0.57 & 0.87 & 0.61 \\
  \toprule
\end{tabular}
}
\end{table}

\section{Conclusions}
\label{sec:typestyle}

In this study, we fine-tuned Large Multimodal Model (LMMs) for Automatic Pronunciation Assessment (APA) using Speechocean762 and a private dataset across multiple granularities and aspects. Fine-tuning on single-granularity tasks significantly outperforms zero-shot settings and achieves competitive results, even surpassing some public and commercial systems.

The fine-tuned model performs strongly at the word and sentence levels, with Pearson Correlation Coefficient (PCC) values approaching state-of-the-art APA systems, indicating robust performance at higher granularities. However, phoneme-level assessment remains challenging, revealing limitations of LMMs in extremely fine-grained tasks.

Additionally, although PCC reaches about 0.9, indicating strong linear correlation, Spearman’s rank Correlation Coefficient (SCC) remains around 0.6. This discrepancy suggests that SCC, which measures ordinal consistency without assuming linearity or Gaussian distribution, may be a more appropriate metric for APA as model performance improves. Future work on LLMs should should focus on enhancing phoneme-level modeling and incorporating rank-based metrics such as SCC to better capture the ordinal nature of APA.


\vfill\pagebreak

\bibliographystyle{IEEEbib}
\bibliography{refs}

\begin{thebibliography}{10}

\bibitem{Kheir2023PA}
Yassine Kheir et~al.,
\newblock ``Automatic pronunciation assessment - a review,''
\newblock in {\em Findings of the ACL: EMNLP}, 2023.

\bibitem{Ye2022Approach}
Wenxuan Ye et~al.,
\newblock ``An approach to mispronunciation detection and diagnosis with acoustic, phonetic and linguistic (apl) embeddings,''
\newblock {\em arXiv preprint arXiv:2110.07274}, 2022.

\bibitem{Wu2024Prompting}
Minglin Wu et~al.,
\newblock ``Prompting large language models with mispronunciation detection and diagnosis abilities,''
\newblock in {\em Interspeech}, 2024.

\bibitem{Mao2019OR}
Shaoguang Mao et~al.,
\newblock ``Nn-based ordinal regression for assessing fluency of esl speech,''
\newblock in {\em ICASSP}, 2019.

\bibitem{Lin2020PA}
Binghuai Lin et~al.,
\newblock ``Automatic scoring at multi-granularity for l2 pronunciation,''
\newblock in {\em Interspeech}, 2020.

\bibitem{Gong2022GOPT}
Yuan Gong et~al.,
\newblock ``Transformer-based multi-aspect multi-granularity non-native english speaker pronunciation assessment,''
\newblock in {\em ICASSP}, 2022.

\bibitem{Chao20233MH}
Fu-An Chao et~al.,
\newblock ``A hierarchical context-aware modeling approach for multi-aspect and multi-granular pronunciation assessment,''
\newblock in {\em INTERSPEECH}, 2023.

\bibitem{Fu2024LLM}
Kaiqi Fu et~al.,
\newblock ``Pronunciation assessment with multi-modal large language models,''
\newblock {\em arXiv preprint arXiv:2407.09209}, 2024.

\bibitem{Wang2025Exploring}
Ke~Wang et~al.,
\newblock ``Exploring the potential of large multimodal models as effective alternatives for pronunciation assessment,''
\newblock {\em arXiv preprint arXiv:2503.11229}, 2025.

\bibitem{Witt2000Phone}
S.M Witt and S.J Young,
\newblock ``Phone-level pronunciation scoring and assessment for interactive language learning,''
\newblock {\em Speech Communication}, 2000.

\bibitem{Hu2015Improved}
Wenping Hu et~al.,
\newblock ``Improved mispronunciation detection with deep neural network trained acoustic models and transfer learning based logistic regression classifiers,''
\newblock {\em Speech Communication}, 2015.

\bibitem{Qian2010Capturing}
Xiaojun Qian et~al.,
\newblock ``Capturing l2 segmental mispronunciations with joint-sequence models in computer-aided pronunciation training (capt),''
\newblock in {\em ISCSLP}, 2010.

\bibitem{Li2016Tone}
Wei Li et~al.,
\newblock ``Using tone-based extended recognition network to detect non-native mandarin tone mispronunciations,''
\newblock in {\em APSIPA}, 2016.

\bibitem{Leung2019CNN}
Wai-Kim Leung et~al.,
\newblock ``Cnn-rnn-ctc based end-to-end mispronunciation detection and diagnosis,''
\newblock in {\em ICASSP}, 2019.

\bibitem{Xu2021Explore}
Xiaoshuo Xu et~al.,
\newblock ``Explore wav2vec 2.0 for mispronunciation detection,''
\newblock in {\em Interspeech}, 2021.

\bibitem{Wang2023Assessing}
Zhiyi Wang et~al.,
\newblock ``Assessing phrase break of esl speech with pre-trained language models and large language models,''
\newblock in {\em INTERSPEECH}, 2023.

\bibitem{Cincarek2009PA}
Tobias Cincarek et~al.,
\newblock ``Automatic pronunciation scoring of words and sentences independent from the non-native’s first language,''
\newblock {\em Computer Speech and Language}, 2009.

\bibitem{Do2023Hierarchical}
Heejin Do et~al.,
\newblock ``Hierarchical pronunciation assessment with multi-aspect attention,''
\newblock in {\em ICASSP}, 2023.

\bibitem{Ryu2023Joint}
Hyungshin Ryu et~al.,
\newblock ``A joint model for pronunciation assessment and mispronunciation detection and diagnosis with multi-task learning,''
\newblock in {\em INTERSPEECH}, 2023.

\bibitem{Chao2025HMamba}
Fu-An Chao and Berlin Chen,
\newblock ``Towards efficient and multifaceted computer-assisted pronunciation training leveraging hierarchical selective state space model and decoupled cross-entropy loss,''
\newblock {\em arXiv preprint arXiv:2502.07575}, 2025.

\bibitem{NatarajanBalaji2025Leveraging}
Shankar~Natarajan Balaji et~al.,
\newblock ``Leveraging asr and llms for automated scoring and feedback in children’s spoken language assessments,''
\newblock in {\em SLaTE}, 2025.

\bibitem{Ma2025Assessment}
Rao Ma et~al.,
\newblock ``Assessment of l2 oral proficiency using speech large language models,''
\newblock in {\em Interspeech}, 2025.

\bibitem{Hu2021LoRA}
Edward~J. Hu et~al.,
\newblock ``Lora: Low-rank adaptation of large language models,''
\newblock {\em arXiv preprint arXiv:2106.09685}, 2021.

\bibitem{Ahn2025English}
Taekyung Ahn and Hosung Nam,
\newblock ``English pronunciation evaluation without complex joint training: Lora fine-tuned speech multimodal llm,''
\newblock {\em arXiv preprint arXiv:2509.02915}, 2025.

\bibitem{Chu2024Qwen2Audio}
Yunfei Chu et~al.,
\newblock ``Qwen2-audio technical report,''
\newblock {\em arXiv preprint arXiv:2407.10759}, 2024.

\bibitem{Rafailov2023DPO}
Rafael Rafailov et~al.,
\newblock ``Direct preference optimization: Your language model is secretly a reward model,''
\newblock {\em arXiv preprint arXiv:2305.18290}, 2023.

\bibitem{Meng2024SimPO}
Yu~Meng et~al.,
\newblock ``Simpo: Simple preference optimization with a reference-free reward,''
\newblock in {\em NeurIPS}, 2024.

\bibitem{Zhang2021speechocean762}
Junbo Zhang et~al.,
\newblock ``speechocean762: An open-source non-native english speech corpus for pronunciation assessment,''
\newblock in {\em Interspeech}, 2021.

\end{thebibliography}

\end{document}